\documentclass{article} % For LaTeX2e
\usepackage{nips15submit_e,times}

\usepackage{hyperref}
\usepackage{url}
\usepackage[utf8]{inputenc}
\usepackage{amsmath,amssymb}
\newcommand{\vertiii}[1]{{\left\vert\kern-0.25ex\left\vert\kern-0.25ex\left\vert #1 
    \right\vert\kern-0.25ex\right\vert\kern-0.25ex\right\vert}}
\usepackage{bbm}
\usepackage{graphicx}
\usepackage{subfig}
\usepackage{caption}
\usepackage{algorithm}
\usepackage{algorithmic}
\usepackage{epsfig}
\usepackage{amsthm}
\usepackage{breqn} %break equations
\usepackage{changebar}
\usepackage{epstopdf}
\usepackage{tabularx}
\usepackage{wrapfig}
\usepackage{enumitem}
\usepackage{mathrsfs}
\newtheorem{lemma}{Lemma}
\newtheorem{theorem}{Theorem}

\newtheorem{corollary}{Corollary}
\newtheorem{proposition}{Proposition}
\newtheorem{example}{Example}

\newtheorem{assumption}{Assumption}
\newtheorem{definition}{Definition}

\usepackage{amssymb,amsmath}

\input xy
\xyoption{all}

\begin{document}

\title{Learning Network of Multivariate Hawkes Processes: A Time Series Approach}

\maketitle
\begin{abstract}
Learning the influence structure of multiple time series data is of great interest to many disciplines.
This paper studies the problem of recovering the causal structure in network of multivariate linear Hawkes processes. 
 In such processes, the occurrence of an event in one process affects the probability of occurrence of new events in some other processes. 
 Thus, a natural notion of causality exists between such processes captured by the support of the excitation matrix.
We show that the resulting causal influence network is equivalent to the Directed Information graph (DIG) of the processes, which encodes the causal factorization of the joint distribution of the processes.
 Furthermore, we present an algorithm for learning the support of excitation matrix (or equivalently the DIG). 
 The performance of the algorithm is evaluated on synthesized multivariate Hawkes networks as well as a stock market and MemeTracker real-world dataset.
\end{abstract}

%\vspace{-.3cm}
\section{Introduction}

In many disciplines, including biology, economics, social sciences, and computer science, it is important to learn the structure of interacting networks of stochastic processes. In particular, succinct representation of the causal interactions in the network is of interest.

A lot of studies in the causality fields focus on causal discovery from time series. To find causal relations from time series, one may fit vector autoregressive models on the time series, or more generally, evaluate the causal influences with transfer entropy \cite{schreiber2000measuring} or directed information \cite{directed}. 
This paper considers learning causal structure for a specific type of time series, multivariate linear Hawkes process \cite{hawkes}. Hawkes processes were originally motivated by the quest for good statistical models for earthquake occurrences. Since then, they have been successfully applied to seismology \cite{23}, biology \cite{28}, criminology \cite{21}, computational finance \cite{7, application, muni2011modelling}, etc. 
It is desirable to develop specific causal discovery methods for such processes and study the properties of existing methods in this particular scenario.

 In multivariate or mutually exciting point processes, occurrence of an event (arrival) in one process affects the conditional probability of new occurrences, i.e., the \textit{intensity} function of other processes in the network. 
Such interdependencies between the intensity functions of a linear Hawkes process are modeled as follows: the intensity function of processes $j$ is assumed to be a linear combination of different terms, such that each term captures only the effects of one other process (See Section \ref{sec:haw}).
  Therefore, a natural notion of functional dependence (causality) exists among the processes in the sense that in linear mutually exciting processes, if the coefficient pertaining to the effects of process $i$ is non-zero in the intensity function of process $j$, we know that process $i$ is influencing process $j$. 
 This dependency is captured by the support of the excitation matrix of the network. As a result, estimation of the excitation (kernel) matrix of multivariate processes is crucial both for learning  the structure of their causal network and for other inference tasks and has been the focus of research. 
 For instance, maximum likelihood estimators were proposed for estimating the parameters of excitation matrices with exponential and Laguerre decay in \cite{ML,ML2}. These estimators depend on existence of i.i.d. samples. However, often we do not have access to i.i.d. samples when analyzing time series.
 Second-order statistics of the multivariate Hawkes processes were used to estimate the kernel matrix of a subclass of multivariate Hawkes processes called symmetric Hawkes processes \cite{bacrynon}. Utilizing the branching property of the Hawkes processes, an expectation maximization algorithm was proposed to estimate the excitation matrix in \cite{EM}.

We aim to investigate efficient approaches to estimation of excitation matrix of Hawkes processes from time series that does not require i.i.d. samples and investigate how the concept of causality in such processes is related to other established approaches to analyze causal effects in time series.

\subsection{Summary of Results and Organization}

%{\color{red}
Our contribution in this paper is two fold. First, we prove that for linear multivariate Hawkes processes, the causal relationships implied by the excitation matrix is equivalent to a specific factorization of the joint distribution of the system called {\it minimal generative model}. 
Minimal generative models encode causal dependencies based on a generalized notion of Granger causality, measured by causally conditioned directed information  \cite{equivalence}. 
One significance of this result is that it provides a surrogate  to directed information measure for capturing causal influences for Hawkes processes.
Thus, instead of estimating the directed information, which often requires estimating a high dimensional joint distribution, it suffices to learn the support of the excitation matrix.
Our second contribution is indeed providing an estimation method for learning the support of  excitation matrices with exponential form using second-order statistics of the Hawkes processes.

Our proposed learning approach, in contrast with the previous work \cite{bacrynon,yang2013mixture}, is not limited to symmetric Hawkes processes. In a symmetric Hawkes process, it is assumed that the Laplace transform of the excitation matrix can be factored into product of a diagonal matrix and a constant unitary matrix. Moreover, it is assumed that the expected values of all intensities are the same. 
A numerical method to approximate the excitation matrix from a set of coupled integral equations was recently proposed in \cite{bacry2014second}. Our approach is based on an exact analytical solution to find the excitation matrix. 
Interestingly, the exact approach turns out to be both more robust and less expensive in terms of complexity compared to the numerical method of \cite{bacry2014second}.

The rest of this paper is organized as follows. Background material, some definitions, and the notation are presented in Section \ref{sec:pri}. Specifically, therein, we formally introduce multivariate Hawkes processes and directed information graphs. In Section \ref{sec:eq}, we establish the connection between the excitation matrix and the corresponding DIG. In Section \ref{sec:learn}, we propose an algorithm for learning the excitation matrix or equivalently the DIG of a class of stationary multivariate linear Hawkes processes. Section \ref{sec:exp} illustrates the performance of the proposed algorithm in inferring the causal structure in a network of synthesized mutually exciting linear Hawkes processes and in stock market. Finally, we conclude our work in Section \ref{sec:con}.

%=========================================================================
\section{Preliminary Definitions}\label{sec:pri}
In this Section we review some basic definitions and our notation. We denote random processes by capital letters and a collection of $m$ random processes by $\underline{X}_{[m]}=\{X_{1},...,X_{m}\}$, where $[m]:=\{1,...,m\}$. We denote the $i$th random process at time $t$ by $X_{i}(t)$, the random process $X_{i}$ from time $s$ up to time $t$ by $X_{i,s}^{t}$, and a subset $\mathcal{K}\subseteq [m]$ of random process up to time $t$ by $\underline{X}_{\mathcal{K}}^{t}$. The Laplace transform and Fourier Transform of $X_{i}$ are denoted, respectively by 
\begin{align}\label{lapli}
&{L}[X_{i}](s)=\int_{0}^{\infty}X_{i}(t)e^{-st}dt, \\ \notag
 &\mathcal{F}[X_{i}](\omega)\ =\int_{-\infty}^{\infty}X_{i}(t)e^{-j\omega t}dt,
\end{align}
where $j=\sqrt{-1}$. The convolution between two functions $f$ and $g$ is defined as $f*g(t):=\int_{\mathbb{R}}f(x)g(t-x)dx$.
The joint distribution of processes $\{X_{1}^{n},...,X_{m}^{n}\}$ is represented by $P_{\underline{X}}(n)$.

%========================================================
\subsection{Multivariate Hawkes Processes}\label{sec:haw}

\begin{figure}[t]
\centering
 \includegraphics[width=88mm, height=58mm]{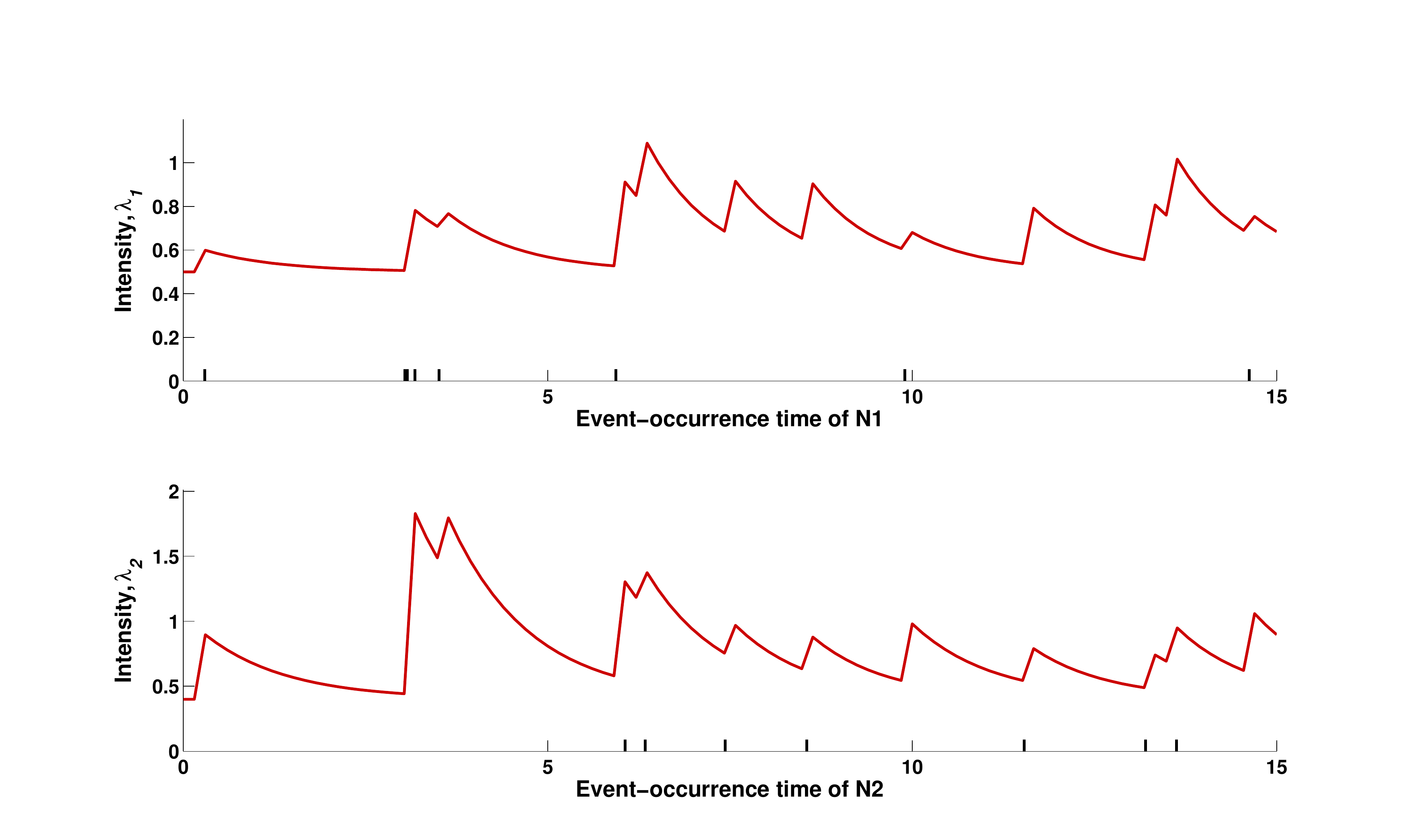}
  \caption{Intensities of the multivariate Hawkes process. }
  \label{Hawke1}
\end{figure}

Fix a complete probability space $(\Omega,\mathcal{F},P)$. Let $N(t)$ denotes the counting process representing  the cumulative number of events up to time $t$ and let  $\{\mathcal{F}^{t}\}_{t\geq0}$ be a set of increasing  $\sigma$-algebras such that $\mathcal{F}^{t}=\sigma\{N^{t}\}$. The non-negative, $\mathcal{F}^{t}$-measurable process $\lambda(t)$ is called the intensity of $N(t)$ if
$$
\small{P(N(t+dt)-N(t)=1|\mathcal{F}^{t})=\lambda(t)dt+o(dt)}.
$$
A classical example of mutually exciting processes, a multivariate Hawkes process \cite{hawkes}, is a multidimensional process $\underline{N}(t)=\{N_{1},...,N_{m}\}$ such that for each $i\in[m]$
\begin{align}\label{prob}
&P\left(dN_{i}(t)=1|\mathcal{\underline{F}}^{t}\right)=\lambda_{i}(t)dt+o(dt), \\ \nonumber
 &P(dN_{i}(t)>1|\mathcal{\underline{F}}^{t})=o(dt),
\end{align}
where $\underline{\mathcal{F}}^{t}=\sigma\{\underline{N}^{t}\}$. The above equations imply that $\mathbb{E}[dN_{i}(t)/dt|\mathcal{\underline{F}}^{t}]=\lambda_{i}(t).$
  Furthermore, the intensities are all positive and are given by
\begin{equation}\label{intensities}
\lambda_{i}(t)=v_{i}+\sum_{k=1}^{m}\int_{0}^{t}\gamma_{i,k}(t-t')dN_{k}(t').
\end{equation}
The exciting functions $\gamma_{i,k}(\cdot)$s are in $\ell_{1}$ such that $\lambda_{i}(t)\geq0$ for all $t>0$. Equivalently, in matrix representation:
\begin{equation}\label{matrix}
\Lambda(t)=\textbf{v}+\int_{0}^{t}\Gamma(t-t')d\underline{N}(t'),
\end{equation}
where $\Gamma(\cdot)$ denotes an $m\times m$ matrix with entries $\gamma_{i,j}(\cdot)$; $d\underline{N}, \Lambda(\cdot)$, and $\textbf{v}$ are $m\times 1$ arrays with entries $dN_{i}, \lambda_{i}(\cdot)$, and $v_{i}$, respectively. Matrix $\Gamma(\cdot)$ is called the excitation (kernel) matrix. Figure \ref{Hawke1} illustrates the intensities of a multivariate Hawkes process comprised of two processes ($m=2$) with the following parameters
$$
\textbf{v}=\begin{pmatrix}
  0.5 \\
  0.4
 \end{pmatrix}\ ,\ \ \Gamma(t)=\begin{pmatrix}
  0.1e^{-t}&0.3e^{-1.1t} \\
  0.5e^{-0.9t}&0.3e^{-t}
 \end{pmatrix}u(t),
$$
where $u(t)$ is the unit step function.

\begin{assumption}\label{strictly}
A joint distribution is called \textit{positive} (non-degenerate), if there exists a reference measure $\phi$ such that $P_{\underline{X}}\ll\phi$ and $\frac{dP_{\small{\underline{\textbf{X}}}}}{d\phi} > 0$, where $P_{\underline{X}}\ll\phi$ denotes that $P_{\underline{X}}$ is absolutely continuous with respect to $\phi$\footnote{A measure $P_{\underline{X}}$ on Borel subsets of the real line is absolutely continuous with respect to measure $\phi$ if for every measurable set $B$, $\phi(B) = 0$ implies $P_{\underline{X}}(B)=0$.}.
\end{assumption}

Note that the Assumption \ref{strictly} states that none of the processes is fully determined by the other processes.

%=========================================================================
\subsection{Causal Structure}\label{sec:causal}

A causal model allows the factorization of the joint distribution in some specific ways.
\textit{Generative model graphs} are a type of graphical model 
that similar to Bayesian networks \cite{pearl} represent a causal factorization of the joint  \cite{directed}.
More precisely, it was shown in \cite{directed} that under Assumption \ref{strictly}, the joint distribution of a causal\footnote{In causal systems, given the full past of the system, the present of the processes become independent.} discrete-time dynamical system with $m$ processes can be factorized as follows,
\begin{equation}\label{abo}
P_{\underline{X}}=\prod_{i=1}^{m}P_{X_{i}||\underline{X}_{B_i}},
\end{equation}
where $B(i)\subseteq-\{i\}$ is the minimal\footnote{Minimal in terms of its cardinality.} set of processes that causes process $X_i$, i.e., parent set of node $i$ in the corresponding minimal generative model graph. Such factorization of the joint distribution is called minimal generative model. In Equation (\ref{abo}),
$$
P_{X_{i}||\underline{X}_{B_i}}:=\prod_{t=1}^nP_{\tiny{X_{i}(t)|\underline{\mathcal{F}}_{B_\cup\{i\}}^{t-1}}},
$$
and 
$
\underline{\mathcal{F}}_{B_\cup\{i\}}^{t-1}=\sigma\{\underline{X}_{{B_\cup\{i\}}}^{t-1}\}
$. 

Extending the definition of generative model graphs to continuous-time systems requires some technicalities which are not necessary for the purpose of this paper. Hence we illustrate the general idea through an example.

The following example demonstrates the minimal generative model graph of a simple continuous-time system.

\begin{example}\label{example12}
Consider a dynamical system in which the processes evolve over time horizon $[0,T]$ through the following coupled differential equations:
\begin{align}\nonumber
&dX_1=f(X_1,X_2)dt+dW,\\ \nonumber
&dX_2=g(X_2)dt+dU, \\ \nonumber
&dX_3=h(X_1,X_2,X_3)dt+dV,
\end{align}
where $W, U$ and $V$ are independent exogenous noises. 
For small time $dt$, this becomes,
\begin{equation}\label{dyn}
\begin{aligned}
&dX_1(t+dt)\approx \Delta f(X_1(t),X_2(t))+dW(t),\\ 
&dX_2(t+dt)\approx \Delta g(X_2(t))+dU(t),\\ 
&dX_3(t+dt)\approx \Delta h(X_1(t),X_2(t),X_3(t))+dV(t).
\end{aligned}
\end{equation}
In this example, since the system is causal, the corresponding joint distribution can be factorized as follows,
\begin{equation}\label{def-first}
P_{\underline{X}}=\prod_{j=1}^{3}\prod_{k\geq0}P_{\tiny{X_{j}(T-kdt)|\underline{\mathcal{F}}^{T-(k+1)dt}}},
\end{equation}
where $\underline{\mathcal{F}}^{T-(k+1)dt}=\sigma\{\underline{X}_{\{1,2,3\}}^{T-(k+1)dt}\}$. 
Due to (\ref{dyn}), we can rewrite (\ref{def-first}) as
\begin{equation}\label{fina}
P_{\underline{X}}=P_{X_{1}||X_{2}}
P_{X_{2}}
P_{X_{3}||X_{1},X_2}.
\end{equation}
Figure \ref{fig:gene} demonstrates the corresponding generative model graph of the factorization in (\ref{fina}). 
\begin{figure}
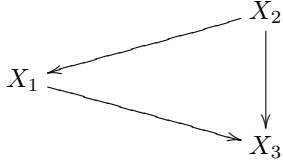

\hspace{4cm}
\xygraph{ !{<0cm,0cm>;<1.9cm,0cm>:<0cm,1.8cm>::} 
!{(-.7,1) }*+{X_1}="x1"
!{(1,1.5) }*+{X_2}="x2"
 !{(1,.5)}*+{X_3}="x3"
"x2":"x1"  "x1":"x3" "x2":"x3" }
   \caption{Minimal generative model graph of Example \ref{example12}.}\label{fig:gene}
\end{figure}
\end{example}

In general, the joint distribution of a causal dynamical system can be factorized as $P_{\underline{X}}=\prod_{i=1}^{m}P_{X_{i}||\underline{X}_{B_i}}$, 
where $B(i)\subseteq-\{i\}$ is the parent set of node $i$ in the corresponding minimal generative model graph, and
$$
P_{X_{i}||\underline{X}_{B_i}}=\prod_{k\geq0}P_{\tiny{X_{i}(T-kdt)|\underline{\mathcal{F}}_{B_i}^{T-(k+1)dt}}}.
$$

%=========================================================================
\section{Two Equivalent Notions of Causality for Multivariate Hawkes Processes}\label{sec:eq}

In linear multivariate Hawkes processes, a natural notion of causation exists in the following sense: if $\gamma_{i,j}\neq0$, then occurrence of an event in $j$th process will affect the likelihood of the arrivals in $i$th process.
Next, we establish the relationship between the excitation matrix of multivariate Hawkes processes and their
generative model graph. To do so, first, we discuss the equivalence of directed information graphs and generative models graphs which was established in \cite{equivalence}.

\subsection{Directed Information Graphs (DIGs)}
An alternative graphical model to encode statistical interdependencies in stochastic causal dynamical systems are \textit{directed information graphs} (DIGs) \cite{directed}.
Such graphs are defined based on an information-theoretic quantity, {\em directed information} (DI) and it was shown in \cite{equivalence} that under some mild assumptions, they are equivalent to the minimal generative model graphs. Hence, DIGs also represent a minimal factorization of the joint distribution.

In a DIG, to determine whether $X_{j}$ causes $X_{i}$ over a time horizon $[0,T]$ in a network of $m$ random processes, two conditional probabilities are compared in KL-divergence sense: one is the conditional probability of $X_i(t+dt)$ given full past, i.e., $\underline{\mathcal{F}}^t:=\sigma\{\underline{X}^t\}$ and the other one is the conditional probability of $X_i(t+dt)$ given full past except the past of $X_j$, i.e., $\underline{\mathcal{F}}^t_{-\{j\}}:=\sigma\{\underline{X}^{t}_{-\{j\}}\}$. 
It is declared that there is no influence from $X_j$ on $X_i$, if the two conditional probabilities are the same.
More precisely, there is an influence from $X_j$ on $X_i$ if and only if the following directed information measure is positive \cite{directed},
\begin{equation}\label{cdif}
\small{
I_{T}(X_{j}\rightarrow X_{i}||\small{\underline{X}}_{-\{i,j\}}):=\inf_{\textbf{t}\in\mathcal{T}(0,T)}\tilde{I}_{\textbf{t}}(X_{j}\rightarrow X_{i}||\small{\underline{X}}_{-\{i,j\}}),
}
\end{equation}
where $-\{i,j\}:=[m]\setminus\{i,j\}$, $\mathcal{T}$ denotes the set of all finite partitions of the time interval $[0,T]$ \cite{weissman2013directed}, and
\begin{align}\notag
 \tilde{I}_{\textbf{t}}(X_{j}\rightarrow X_{i}||\small{\underline{X}}_{-\{i,j\}}):= \sum_{k=0}^nI\left(X_{i,t_{k-1}}^{t_k};X_{j,0}^{t_k}|\mathcal{F}^{t_{k-1}}_{-\{j\}}\right),
\end{align}
where $\textbf{t}:=(0=t_0,t_1,...,t_{n}=T)$.
Finally, $I(X;Y|Z)$ represents the conditional mutual information between $X$ and $X$ given $Z$ and it is given by
\begin{align} \label{mutual}\notag
I(X;Y|Z):=\mathbb{E}_{P_{X,Y,Z}}\left[\log \frac{dP_{X|Y,Z}}{dP_{X|Z}}\right].
\end{align}

\subsection{Equivalence between Generative Model Graph and Support of Excitation Matrix} 

As mentioned earlier, the corresponding minimal generative model graph and the DIG of a causal dynamical system are equivalent. Thus, to characterize the corresponding minimal generative model graphs of a multivariate Hawkes system, we study the properties of its corresponding DIG.

\begin{proposition}\label{l1}
Consider a set of mutually exciting processes $\underline{N}$ with excitation matrix $\Gamma(t)$. Under Assumption \ref{strictly}, $I_T(N_{j}\rightarrow N_{i}||{\underline N}_{-\{i,j\}})=0$ if and only if $\gamma_{i,j}\equiv0$ over time interval $[0,T]$.
\end{proposition}
\textbf{Proof:}\ See Section \ref{l1p}. $\square$

Proposition \ref{l1} signifies that the support of the excitation matrix $\Gamma(\cdot)$ determines the adjacency matrix of the DIG and vice versa. Therefore, learning DIG of a mutually exciting Hawkes processes satisfying Assumption \ref{strictly} is equivalent to learning the excitation matrix given samples from each of the processes.
In other word, in the presence of side information that the processes are Hawkes, it is more efficient to learn the causal structure through learning the excitation matrix rather than the directed information needed for learning the DIG in general.

%=========================================================================

\section{Learning the Excitation Matrix}\label{sec:learn}
In this section, we present an approach for learning the causal structure of a stationary Hawkes network with exponential exciting functions through learning the excitation matrix. This method is based on second order statistic of the Hawkes processes and it is suitable for the case when no i.i.d. samples are available. Note that when i.i.d. samples are available, non-parametric methods for learning the excitation matrix such as MMEL algorithm \cite{ML2} exist. In this approach the exciting functions are expressed as linear combination of a set of base kernels and a penalized likelihood is used to estimate the parameters of the model. 
As mentioned earlier, we focus on learning the excitation matrix of multivariate Hawkes processes with exponential exciting functions. This class of Hawkes processes has been widely applied in many areas such as seismology, criminology, and finance \cite{23,28,21,7}. 
\begin{definition}
The excitation matrix of a multivariate Hawkes processes with exponential exciting functions is defined as follows
\begin{align}\nonumber
&\!\!\!\!\!\mathcal{E}\textit{xp}(m):=\{\sum_{d=1}^{D}A_{d}e^{-\beta_{d}t}u(t): A_{d}\in\mathbb{R}^{m\times m}, \\
&\:\:\:\; \vspace{-1mm}\tiny{(\sum_{d=1}^{D}A_{d}e^{-\beta_{d}t})_{i,j}\geq0, \rho(\sum_{d=1}^{D}\frac{A_{d}}{\beta_{d}})<1, D\in\mathbb{N} \},}
\end{align}
where $\{\beta_{d}\}>0$ is called the set of exciting modes. 
\end{definition}
\begin{example}\label{example1}
Consider a set of $m=5$ mutually exciting processes with the following exponential excitation matrix 

\begin{align}\nonumber
&\begin{scriptsize}
\begin{pmatrix}
  2& 0 & 0 & 0 &0\\
  0 & 0& .5 & 0& 0\\
  0 & 1.5 & 0 & 0&0\\
 0& 0 & 0 & 1.3 & 0\\
  0 &0 & 0 &0 & 1
 \end{pmatrix}  \frac{e^{-t}}{20} \end{scriptsize}
  \begin{scriptsize}
 +\begin{pmatrix}
  0& 0& .5& 0 &0\\
  0 & 0& 0 & 0& 2\\
  0 & 1 & 0 & 2.5 &0\\
  .1 & 0 & 0 & 0 & 0\\
  0 &0 & 0 &1 & 0
 \end{pmatrix}\frac{e^{-1.4t}}{20} \end{scriptsize}
 \\ \label{simuf1}
& \begin{scriptsize}
  +
\begin{pmatrix}
  1& 1.5 & 1 & 0 &0\\
  0 & 0& 0 & 0& -1\\
  0 & 0 & 2 & 0 &0\\
  2 & 0 & 0 & 0 & 0\\
  0 &0 & 0 & 0 & 0
 \end{pmatrix} 
\frac{e^{-2t}}{20}  \end{scriptsize}
\end{align}
In this example $D=3$ and the exciting modes are $\{1,1.4,2\}$. By Proposition \ref{l1}, the adjacency matrix of the corresponding DIG of this network is given by the support of its excitation matrix. Figure \ref{simu123} depicts the corresponding DIG.

\begin{figure}
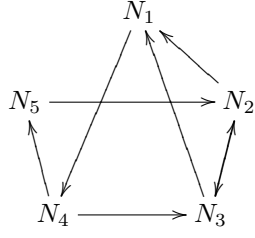

\hspace{5cm} 
\xygraph{ !{<0cm,0cm>;<1.9cm,0cm>:<0cm,1.5cm>::} 
!{(-.2,.8) }*+{N_{1}}="x1"
 !{(-1,0) }*+{N_{5}}="x5"
!{(.5,0) }*+{N_{2}}="x2"
 !{(-.8,-1)}*+{N_{4}}="x4"
 !{(.3,-1)}*+{N_{3}}="x3"
"x2":"x1" "x3":"x1" "x3":"x2" "x5":"x2"  "x2":"x3" "x4":"x3"  "x1":"x4"  "x4":"x5"}
\caption{Corresponding DIG of the network in Example \ref{example1} with the excitation matrix given by (\ref{simuf1}) }\label{simu123}
\end{figure}

\end{example}

Before describing our algorithm, we need to derive some useful properties of moments of the process. A multivariate Hawkes process with the excitation matrix $\Gamma$ has stationary increments, i.e., the intensity processes is stationary, if and only if the following assumption holds \cite{hawkes,stationaryp}:

\begin{assumption}\label{ass2}
The spectral radius (the supremum of the absolute values of the eigenvalues) of the matrix $\overline{\Gamma}$, where $[\overline{\Gamma}]_{i,j}=||\gamma_{i,j}||_{1}$ is strictly less than one, i.e., $\rho(\overline{\Gamma})<1$.
\end{assumption}
In this case, from (\ref{matrix}) and Equation (\ref{prob}):

\begin{align}\nonumber
&\Lambda=\mathbb{E}[\Lambda(t)]=\textbf{v}+\int_{0}^{t}\Gamma(t-t')\mathbb{E}[d\underline{N}(t')]\\ \label{mean}
&=\textbf{v}+\int_{0}^{t}\Gamma(t-t')\Lambda dt'=\textbf{v}+\overline{\Gamma}\Lambda.
\end{align}
By Assumption 2, $\sum_{i\geq0}\overline{\Gamma}^{i}$ converges to $(I-\overline{\Gamma})^{-1}$, thus 
$\Lambda=(I-\overline{\Gamma})^{-1}\textbf{v}$.
 The normalized covariance matrix of a stationary multivariate Hawkes process with lag $\tau$ and window size $z>0$ is defined by
\begin{equation}\label{cov}
\Sigma_{z}(\tau):=\frac{1}{z}\mathbb{E}\left[\int_{t}^{t+z}d\underline{N}(x)\int_{t+\tau}^{t+\tau+z}(d\underline{N}(y))^{T}\right]-\Lambda\Lambda^{T}z,
\end{equation}
where $\int_{t}^{t+t'}d\underline{N}(x)$ denotes the number of events in time interval $(t,t+t']$.

\begin{theorem}\cite{bacrynon}
The Fourier transform of the normalized covariance matrix of a stationary multivariate Hawkes process with lag $\tau$ and window size $z>0$ is given by 
\begin{align}\label{lapl}
&\mathcal{F}[\Sigma_{z}](-\omega)\\  \nonumber
&\begin{scriptsize}
=4\frac{\sin^{2}{z\omega/2}}{\omega^{2}z}\left(I-\mathcal{F}[\Gamma](\omega)\right)^{-1}diag(\Lambda)\left(I-\mathcal{F}[\Gamma](\omega)\right)^{-\dagger},
\end{scriptsize}
 \end{align}
where $A^{\dagger}$ denotes the Hermitian conjugate of matrix $A$, and $diag(\Lambda)$ is a diagonal matrix with vector $\Lambda$ as the main diagonal.
\end{theorem}

In order to learn the excitation matrix with exponential exciting functions, we need to learn the exciting modes $\{\beta_{d}\}$, the number of components $D$, and coefficient matrices $\{A_{d}\}$. Next results establishes the relationship between the exciting modes and the number of components $D$ with the normalized covariance matrix of the process.

\begin{corollary}\label{col}
Consider a network of a stationary multivariate Hawkes processes with excitation matrix $\Gamma(t)$ belonging to $\mathcal{E}\textit{xp}(m)$. Then the exciting modes of $\Gamma(t)$ are the absolute values of the zeros of  $1/ Tr\mathcal{F}[\Sigma_z]^{-1}(\omega)$.
\end{corollary}

\textbf{Proof:}\ See Section \ref{colp}. $\square$

Next, we need to find the coefficient matrices $\{A_d\}$. To do so, we use the covariance density of the processes. The covariance density of a stationary multivariate Hawkes process for $\tau>0$ is defined as \cite{hawkes} 
\begin{equation}\label{est1}
\Omega(\tau):=\mathbb{E}\left[ (d\underline{N}(t+\tau)/dt-\Lambda)(d\underline{N}(t)/dt-\Lambda)^{T}\right].
\end{equation}
Since the processes have stationary increments, we have $\Omega(-\tau)=\Omega^{T}(\tau)$. 

\begin{lemma}\cite{hawkes}\label{l22}
\begin{align}\label{covariance}
\Omega(\tau)= \Gamma(\tau) diag(\Lambda)+\Gamma*\Omega(\tau), \tau>0.
\end{align}
\end{lemma}

It has been shown in \cite{bacry2014second} that the above equation admit a unique solution for $\Gamma(\tau)$.
Next proposition provides a system of linear equations that allows us to learn the coefficient matrices.
%It has been shown in \cite{bacry} that
%\begin{align}\label{covar}\nonumber
%\Sigma_{z}(\tau)=g_{z}(\tau)diag(\Lambda)+g_{z}*\Psi(-\tau)diag(\Lambda)\\ \nonumber
%+g_{z}*diag(\Lambda)\Psi^{T}(\tau)+g_{z}*\widehat{\Psi}*diag(\Lambda)\Psi^{T}(\tau),
%\end{align}

%where $\widehat{\Psi}(t)=\Psi(-t)$, $g_{z}(t)=\max\{(1-\frac{|t|}{z}),0\}$, and $\Psi(t)=\sum_{n\geq1}\Gamma^{(*n)}(t)$ in which $\Gamma^{(*n)}(t)$ denotes the $n$th auto-convolution of $\Gamma(t)$. It is more convenient to represent $\Omega(\tau)$ and $\Sigma_z(\tau)$ in the Laplace and the Fourier domain, respectively:
\begin{proposition}\label{pro}
Consider a network of a stationary multivariate Hawkes processes with excitation matrix $\Gamma(t)\in\mathcal{E}\textit{xp}(m)$, and exciting modes $\{\beta_1,...,\beta_D\}$. Then $\{A_d\}$ are a solution of the linear system of equations: $\textbf{S}=\textbf{A}\textbf{H}$, where $\textbf{H}_{m^2\times m^2}$ is a block matrix with $(i,j)$th block given by
$$
\textbf{H}_{i,j}=\frac{diag(\Lambda)+\mathcal{L}[\Omega](\beta_{j})+\mathcal{L}[\Omega]^{T}(\beta_{i})}{\beta_{j}+\beta_{i}}, 
$$
and $\textbf{A}=\left[A_{1}, ..., A_{D}\right]$ and $\textbf{S}=\left[\mathcal{L}[\Omega](\beta_{1}), ..., \mathcal{L}[\Omega](\beta_{D})\right]$.
%Note that $\mathcal{F}[g_{z}](\omega)=4\frac{\sin^{2}{z\omega/2}}{\omega^{2}z}$ and because of $\mathcal{F}[\Gamma^{(*n)}]=\mathcal{F}[\Gamma]^{n}$ and Assumption \ref{ass2}, we have $I+\mathcal{F}[\Psi]=(I-\mathcal{F}[\Gamma])^{-1}$.
\end{proposition}
\textbf{Proof:}\ See Section \ref{prop}.$\square$

Combining the results of Corollary \ref{col} and Proposition \ref{pro} allows us to learn the excitation matrix of exponential multivariate Hawkes processes from the second order moments. Consequently applying  Proposition \ref{l1}, the causal structure of the network can be learned by drawing an arrow from node $i$ to $j$, when  $\sum_{d=1}^{D}|(A_{d})_{j,i}|>0$.

\subsection{Estimation and Algorithm}
This section discusses estimators for the second order moments, namely the normalized covariance matrix and the covariance density of a stationary multivariate Hawkes processes from data. Once such estimators are available, the approach of previous section maybe used to learn the network. 
The most intuitive estimator for $\Lambda$ defined by Equation (\ref{mean}) is $\underline{N}(T)/T$. It turns out that this estimator converges almost surely to $\Lambda$ as $T$ goes to infinity \cite{bacry}. Furthermore, \cite{bacry} proposes an empirical estimator for the normalized covariance matrix as follows 
\begin{equation}\label{est22}
\small{\widehat{\Sigma}_{z,T}(\tau):=\frac{1}{T}\sum_{i=1}^{\lfloor T/z\rfloor}(X_{iz}-X_{(i-1)z})(X_{iz+\tau}-X_{(i-1)z+\tau})^{T},}
\end{equation}
where $X_{t}:=\underline{N}(t)-\Lambda t$. In the same paper, it has been shown that under Assumption 2, the above estimator converges in $\ell_{2}$ to the normalized covariance matrix (\ref{cov}), i.e., $\widehat{\Sigma}_{z,T}(\tau)\underset{T\rightarrow\infty}{\longrightarrow}\Sigma_{z}(\tau)$.
Notice that the normalized covariance matrix and the covariance density are related by $\Sigma_{dt}(\tau)/dt=\Omega^{T}(\tau)$. Therefore, we can estimate the covariance density matrix using Equation (\ref{est22}) by choosing small enough window size $z=\Delta$. Namely, $\small{\widehat{\Omega}_{\Delta}^T(\tau)=\widehat{\Sigma}_{\Delta}(\tau)/\Delta}$. 
 \begin{algorithm}[H]
  \caption{}
  \label{algorithm}
    \begin{algorithmic}[1]
       \STATE $Input:\ \ \underline{N}^{T}$.
    \STATE $Output:$\ \ DIG.
    \STATE  $\widehat{\Lambda}\leftarrow \underline{N}(T)/T$
    \STATE {Choose $\sigma>0$, $z>0$, and small $\Delta>0$.}     
     \STATE Compute $\widehat{\Sigma}_{z,T}(\tau)$ and $\widehat{\Omega}_{\Delta}(\tau)$ using (\ref{est22}).
     \STATE  $\{\widehat{\beta}_{d}\}_{d=1}^{\widehat{D}}\leftarrow$ Zeros of $1/ Tr\mathcal{F}[\Sigma_z]^{-1}(\omega)$.
     \STATE Compute $\mathcal{L}[\widehat{\Omega}_{\Delta}](\widehat{\beta}_{d})$ for $d=1,...,\widehat{D}$.
     \STATE Solve the set of equations arises from (\ref{dasgha}) for $\widehat{A}_{d}$.
     \STATE Draw $(j,i)$ if $\sum_{d=1}^{\widehat{D}}|(\widehat{A}_{d})_{i,j}|\geq\sigma$.
   \end{algorithmic}
\end{algorithm}
Algorithm \ref{algorithm} summarizes the steps of our proposed approach for learning the excitation matrix and consequently the causal structure of an exponential multivariate Hawkes process.

\section{Experimental Results}\label{sec:exp}
In this section, we present our experimental results for both synthetic and real data.

\subsection{Synthetic Data}
\begin{figure*}[t]
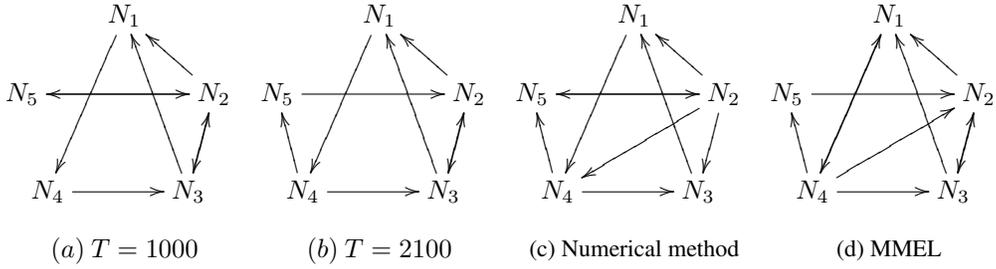

\hspace{.1cm}
  \xygraph{ !{<0cm,0cm>;<1.7cm,0cm>:<0cm,1.3cm>::} !{(-.2,.8) }*+{N_{1}}="x1"
  !{(-1,0) }*+{N_{5}}="x5"
!{(.5,0) }*+{N_{2}}="x2"
 !{(-.8,-1)}*+{N_{4}}="x4"
 !{(.3,-1)}*+{N_{3}}="x3"
   !{(-.2,-1.6)}*+{\small{(a)\ T=1000}}
"x2":"x1" "x3":"x1" "x3":"x2" "x5":"x2"  "x2":"x3" "x4":"x3"   "x2":"x5"  "x1":"x4" } 
\hspace{.1cm}  \xygraph{ !{<0cm,0cm>;<1.7cm,0cm>:<0cm,1.3cm>::} !{(-.2,.8) }*+{N_{1}}="x1"
 !{(-1,0) }*+{N_{5}}="x5"
!{(.5,0) }*+{N_{2}}="x2"
 !{(-.8,-1)}*+{N_{4}}="x4"
 !{(.3,-1)}*+{N_{3}}="x3"
   !{(-.2,-1.6)}*+{\small{(b)\ T=2100}}
"x2":"x1" "x3":"x1" "x3":"x2" "x5":"x2"  "x2":"x3" "x4":"x3"    "x1":"x4" "x4":"x5"}
\hspace{.1cm} \xygraph{ !{<0cm,0cm>;<1.7cm,-0cm>:<0cm,1.3cm>::} !{(-.2,.8) }*+{N_{1}}="x1"
 !{(-1,0) }*+{N_{5}}="x5"
!{(.5,0) }*+{N_{2}}="x2"
 !{(-.8,-1)}*+{N_{4}}="x4"
 !{(.3,-1)}*+{N_{3}}="x3"
  !{(-.2,-1.6)}*+{\small{\text{(c) Numerical method}}}
"x2":"x1" "x3":"x1"  "x5":"x2"  "x2":"x3"  "x2":"x5" "x4":"x5" "x4":"x3" "x1":"x4" "x2":"x4"} 
\hspace{.1cm}  \xygraph{ !{<0cm,0cm>;<1.7cm,0cm>:<0cm,1.3cm>::} 
!{(-.2,.8) }*+{N_{1}}="x1"
 !{(-1,0) }*+{N_{5}}="x5"
!{(.5,0) }*+{N_{2}}="x2"
 !{(-.8,-1)}*+{N_{4}}="x4"
 !{(.3,-1)}*+{N_{3}}="x3"
   !{(-.2,-1.6)}*+{\small{\text{(d) MMEL}}}
"x2":"x1" "x3":"x1"  "x4":"x1"   "x3":"x2" "x2":"x3" "x4":"x3"  "x5":"x2"  "x1":"x4" "x4":"x5"  "x4":"x2"}
  \caption{Recovered DIG of the network in Example \ref{example1} with the excitation matrix given by (\ref{simuf1}), (a), (b) Algorithm 1 with $\Delta=0.2$, $z=2$, and $\small{T\in\{ 1000, 2100\}}$, (c) the numerical method of \cite{bacry2014second} with $Q=70$ and $T=2100$, and (d) MMEL with $35$ i.i.d. samples each of length $60$. Our approach learns the graph with $T=2100$, while other approaches fail at the same sample size.}
  \label{simu1}
\end{figure*}

We apply the proposed algorithms to learn the causal structure of the multivariate Hawkes network of Example \ref{example1} with $\textbf{v}=(0.5, 0.4, 0.5, 1, 0.3)^{T}$. This network satisfies Assumption 2, since $\small{\rho(\overline{\Gamma})\approx0.16}$. The exciting modes are $\{1, 1.4, 2\}$. We observed the arrivals of all processes during a time period $T$. Figure \ref{simu1} depicts the outputs of algorithms 1 for $\Delta=0.2$, $z=2$, and observation lengths $\small{T\in\{1000,2100\}}$. As illustrated in Figure \ref{simu1}, by increasing the length of observation $T$, the output graph converges the true DIG shown in Figure \ref{simu123}. As a comparison, we applied the MMEL algorithm proposed in \cite{ML2} to learn the excitation matrix for this example and the numerical method based on Nystrom method proposed in \cite{bacry2014second} with $T=2100$ and the number of quadrature $Q=70$. Since MMEL requires i.i.d. samples, we generate $35$ i.i.d. samples each of length $60$ to obtain Figure \ref{simu1}(MMEL). Our proposed algorithm outperforms both MMEL and the numerical method of  \cite{bacry2014second}.

\begin{figure}[t]
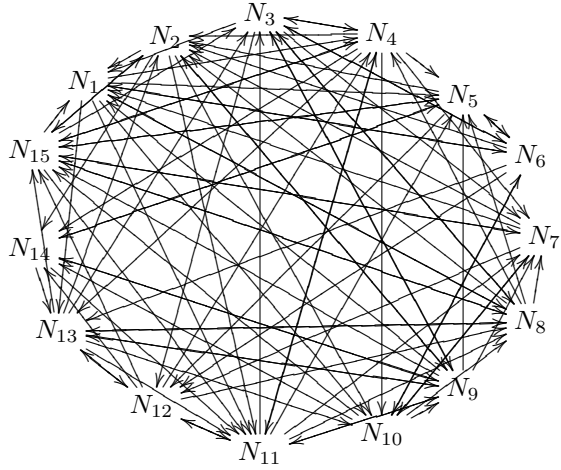

 \vspace{0cm}
 \hspace{4cm}
\xygraph{ !{<0cm,0cm>;<1.8cm,-.3cm>:<-0cm,1.3cm>::} 
 !{(-.8,0.5) }*+{N_1}="x1"
!{(-.2,1.1) }*+{N_2}="x2"
 !{(.5,1.5) }*+{N_3}="x3" 
 !{(1.4,1.5)}*+{N_4}="x4"
 !{(2,1) }*+{N_5}="x5"  
!{(2.5,.5) }*+{N_6}="x6"
!{(2.6,-.3) }*+{N_7}="x7" 
!{(2.5,-1.2) }*+{N_8}="x8" 
!{(2,-2.) }*+{N_9}="x9" 
!{(1.4,-2.6) }*+{N_{10}}="x10" 
!{(.5,-3) }*+{N_{11}}="x11" 
!{(-.3,-2.7) }*+{N_{12}}="x12" 
!{(-1,-2.1) }*+{N_{13}}="x13" 
!{(-1.2,-1.3) }*+{N_{14}}="x14" 
!{(-1.2,-.3) }*+{N_{15}}="x15" 
%!{(.8,-3.5) }*+{\text{(a) True causal networks.}} 
"x1":"x4""x1":"x5""x1":"x6""x1":"x8""x1":"x10""x1":"x11""x1":"x13"
"x2":"x1" "x2":"x9" "x2":"x10""x2":"x11""x2":"x14""x2":"x15"
"x3":"x1""x3":"x4""x3":"x6""x3":"x7""x3":"x8""x3":"x9""x3":"x12"
"x4":"x2""x4":"x3""x4":"x5""x4":"x7""x4":"x10""x4":"x11""x4":"x13""x4":"x14""x4":"x15"
"x5":"x2""x5":"x6""x5":"x12""x5":"x14""x5":"x15"
"x6":"x1""x6":"x2""x6":"x3""x6":"x5""x6":"x10""x6":"x12""x6":"x13"
"x7":"x1""x7":"x2""x7":"x10""x7":"x12""x7":"x15"
"x8":"x1""x8":"x2""x8":"x3""x8":"x4""x8":"x5""x8":"x7""x8":"x13""x8":"x15"
"x9":"x1""x9":"x2""x9":"x3""x9":"x5""x9":"x7""x9":"x11""x9":"x13""x9":"x14""x9":"x15"
"x10":"x6""x10":"x7""x10":"x8""x10":"x9""x10":"x14""x10":"x15"
"x11":"x3" "x11":"x4""x11":"x5""x11":"x7""x11":"x8""x11":"x9""x11":"x12"
"x12":"x4""x12":"x8""x12":"x11""x12":"x13""x12":"x14""x12":"x15"
"x13":"x2""x13":"x3""x13":"x8""x13":"x9""x13":"x10""x13":"x11""x13":"x12""x13":"x15"
"x14":"x5""x14":"x9""x14":"x11""x14":"x13"
"x15":"x1""x15":"x2""x15":"x3""x15":"x4""x15":"x5""x15":"x7""x15":"x8""x15":"x11"}
\caption{True causal structure of the synthesized example.}  \label{anth}
\end{figure}

Furthermore, we conducted another experiment for a network of $15$ processes with $102$ edges illustrated in Figure \ref{anth}. For a sample of length $T=2500$, our algorithm was able to recover $70$ edges correctly but identified $34$ false arrows. MMEL could only recover $58$ arrows correctly while detecting another $41$ false arrows. The input for MMEL was $25$ sequences each of length $100$. %Figure \ref{True} illustrates the true network while figures \ref{1lg1} and \ref{mmel} demonstrate the outcome of our algorithm and MMEL, respectively.

 \begin{figure*}[t]
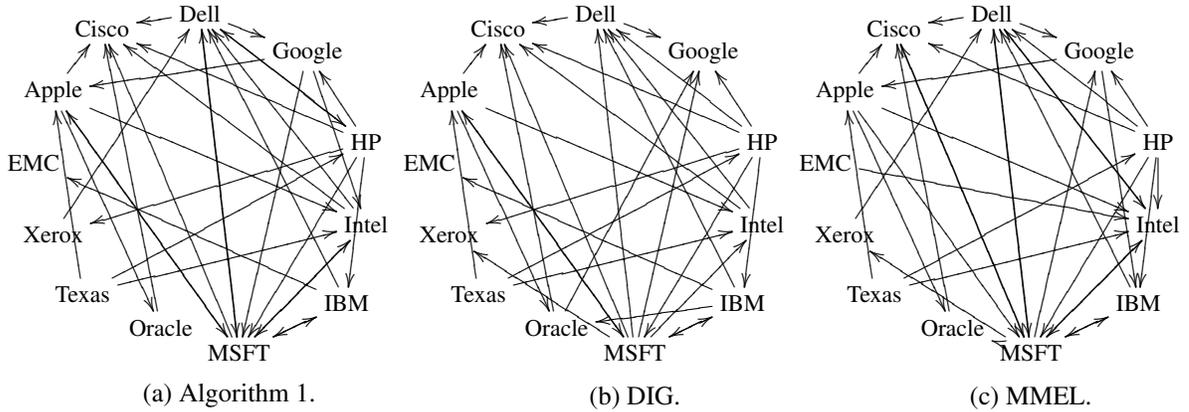

 \hspace{-.5cm}
\xygraph{ !{<0cm,0cm>;<1.3cm,-.3cm>:<-0cm,1cm>::} 
 !{(-1,0) }*+{\small{\text{Apple}}}="x1"
!{(-.5,1) }*+{\small{\text{Cisco}}}="x2"
 !{(.5,1.5) }*+{\small{\text{Dell}}}="x3" 
 !{(1.6,1.3)}*+{\small{\text{Google}}}="x5"
 !{(2.2,.3) }*+{\small{\text{HP}}}="x6"  
!{(2.2,-.8) }*+{\small{\text{Intel}}}="x7"
!{(2,-1.9) }*+{\small{\text{IBM}}}="x10" 
!{(.9,-2.9) }*+{\small{\text{MSFT}}}="x8" 
!{(.1,-2.8) }*+{\small{\text{Oracle}}}="x9" 
!{(-.7,-2.6) }*+{\small{\text{Texas}}}="x11" 
!{(-1,-1.9) }*+{\small{\text{Xerox}}}="x12" 
!{(-1.2,-1) }*+{\small{\text{EMC}}}="x4" 
!{(.8,-3.5) }*+{\text{(a) Algorithm 1.}} 
"x1":"x2"
"x1":"x8"
"x1":"x9"
"x1":"x7"
"x3":"x2"
"x3":"x5"
"x11":"x1"
"x3":"x6"
"x3":"x8"
"x5":"x1"
"x5":"x7"
"x5":"x8"
"x6":"x10"
"x6":"x2"
"x6":"x3"
"x6":"x5"
"x6":"x8"
"x6":"x12"
"x7":"x2"
"x7":"x3"
"x7":"x8"
"x8":"x1"
"x8":"x3"
"x8":"x2"
"x8":"x7"
"x8":"x10"
"x9":"x2"
"x10":"x3"
"x10":"x8"
"x10":"x4"
"x11":"x6"
"x11":"x7"
"x12":"x3"
} 
\xygraph{ !{<0cm,0cm>;<1.3cm,-.3cm>:<0cm,1cm>::} 
 !{(-1,0) }*+{\small{\text{Apple}}}="x1"
!{(-.5,1) }*+{\small{\text{Cisco}}}="x2"
 !{(.5,1.5) }*+{\small{\text{Dell}}}="x3" 
 !{(1.6,1.3)}*+{\small{\text{Google}}}="x4"
 !{(2.2,.3) }*+{\small{\text{HP}}}="x5"  
!{(2.2,-.8) }*+{\small{\text{Intel}}}="x6"
!{(2,-1.9) }*+{\small{\text{IBM}}}="x7" 
!{(.9,-2.9) }*+{\small{\text{MSFT}}}="x8" 
!{(.1,-2.8) }*+{\small{\text{Oracle}}}="x9" 
!{(-.7,-2.6) }*+{\small{\text{Texas}}}="x10" 
!{(-1,-1.9) }*+{\small{\text{Xerox}}}="x11" 
!{(-1.2,-1) }*+{\small{\text{EMC}}}="x12" 
!{(.9,-3.5) }*+{\text{(b) DIG.}}
 "x1":"x2"
"x1":"x6"
"x1":"x8"
"x1":"x9"
"x3":"x2"
"x3":"x4"
"x5":"x2"
"x5":"x3"
"x5":"x4"
"x5":"x7"
"x5":"x8"
"x5":"x11"
 "x6":"x2"
 "x6":"x3"
 "x7":"x9"
  "x7":"x8"
   "x7":"x12"
   "x7":"x3"
   "x8":"x11"
  "x8":"x1"
  "x8":"x2"
  "x8":"x3"
  "x8":"x4"
  "x8":"x6"
  "x8":"x7"
  "x9":"x2"
  "x9":"x4"
  "x10":"x1"
  "x10":"x5"
  "x10":"x6"}
  \xygraph{ !{<0cm,0cm>;<1.3cm,-.3cm>:<0cm,1cm>::} 
 !{(-1,0) }*+{\small{\text{Apple}}}="x1"
!{(-.5,1) }*+{\small{\text{Cisco}}}="x2"
 !{(.5,1.5) }*+{\small{\text{Dell}}}="x3" 
 !{(1.6,1.3)}*+{\small{\text{Google}}}="x4"
 !{(2.2,.3) }*+{\small{\text{HP}}}="x5"  
!{(2.2,-.8) }*+{\small{\text{Intel}}}="x6"
!{(2,-1.9) }*+{\small{\text{IBM}}}="x7" 
!{(.9,-2.9) }*+{\small{\text{MSFT}}}="x8" 
!{(.1,-2.8) }*+{\small{\text{Oracle}}}="x9" 
!{(-.7,-2.6) }*+{\small{\text{Texas}}}="x10" 
!{(-1,-1.9) }*+{\small{\text{Xerox}}}="x11" 
!{(-1.2,-1) }*+{\small{\text{EMC}}}="x12" 
!{(.9,-3.5) }*+{\text{(c) MMEL.}}
"x4":"x1" 
"x10":"x1"
"x1":"x2"
"x8":"x2"
"x5":"x2"
"x9":"x2"
"x3":"x2"
"x5":"x3"
"x11":"x3"
"x8":"x3"
"x3":"x4"
"x5":"x4"
"x8":"x4"
"x10":"x5"
"x1":"x6"
"x8":"x6"
"x3":"x6"
"x5":"x6"
"x12":"x6"
"x10":"x6"
"x1":"x8"
"x2":"x8"
"x3":"x8"
"x5":"x8"
"x7":"x8"
"x7":"x3"
"x6":"x8"
"x6":"x3"
"x8":"x9"
"x1":"x9"
"x4":"x7"
"x5":"x7"
"x8":"x7"
"x8":"x11"
}
\caption{Causal structures for the S\&P (a) using Algorithm \ref{algorithm}, (b) by estimating the directed information DIG, and (c) using MMEL algorithm.}  \label{algco1}
\end{figure*}

\subsection{Stock Market Data }
As an example of how our approach may discover causal structure in real-world data, we analyzed the causal relationship between stock prices of 12 technology companies of the New York Stock Exchange sourced from Google Finance. The prices were sampled every 2 minutes for twenty market days (03/03/2008 - 03/28/2008). Every time a stock price changed by $\pm1\%$ of its current price an event was logged on the stock's process. In order to prevent the substantial changes in stock's prices due to the opening and closing of the market, we ignored the samples at the beginning and at the end of each working day. For this part, we have assumed that the jumps occurring in stock's prices are correlated through a multivariate Hawkes process. This model class was advocated in \cite{linderman2014discovering, bacry}. Figure \ref{algco1}(a) illustrate the causal graph resulting from Algorithm \ref{algorithm}, with $z=30$ and $\Delta=2$ minutes.
 
To compare our learning approach with other approaches, we applied the MMEL algorithm to learn the corresponding causal graph. For this scenario, we assumed that the data collected from each day is generated i.i.d. Hence, a total of $20$ i.i.d. samples were used. Figure \ref{algco1}(c) illustrates the resulting graph.
As one can see, Figures \ref{algco1}(a) and \ref{algco1}(c) convey pretty much a similar causal interactions in the dataset. For instance both of these graphs suggest that one of the most influential companies in that period of time was Hewlett-Packard (HP). Looking into the global PC market share during 2008, we find that this was indeed the case.\footnote{Gartner, \tt\scriptsize http://www.gartner.com/newsroom/id/856712} 

To use another modality, we derive the corresponding DIG of this network applying Equation (\ref{cdif}). For this part, we used the market based on the Black-Scholes model \cite{black1973pricing} in which the stock's prices are modeled via a set of coupled stochastic PDEs. We assumed that the logarithm of the stock's prices are jointly Gaussian and therefore the corresponding DIs were estimated using Equation (24) in \cite{acc2014}.
The resulting DIG is shown in Figure \ref{algco1}(b). Note that this DIG is derived from the logarithm of prices and not the jump processes we used earlier. Still it shares a lot of similarities with the two other graphs. For instance, it also identifies HP as one of the most influential companies and Microsoft as one the most influenced companies in that time period. 

\begin{small}
\begin{center}
  \begin{tabular}{ | l | l | l | p{.9cm} |}
    \hline
     & Alg. 1 & DIG & MMEL \\ \hline
    Alg. 1 & 33 & 25 & 26 \\ \hline
    DIG & 25 & 30 & 24 \\ \hline
    MMEL & 26 & 24 & 34 \\
    \hline
    \end{tabular}
\label{res}
\end{center}
\end{small}

This table shows the number of edges that each of the above approaches recovers and the number of edges that they jointly recover.
This demonstrates the power of exponential kernels even when data does not come from such a model class.

%\cite{marvcek1998stock}

\subsection{MemeTracker Data}
We also studied causal influences in a blogosphere. The causal flow of information between media sites may be captured by studying hyperlinks provided in one media site to others. Specifically, the time of such linking can be modeled using a linear multivariate Hawkes processes with exponential exciting functions \cite{ML2,pinto2015trend}.
This model is also intuitive in the sense that after emerging a new hot topic, in the first several days, the blogs or websites are more likely feature that topics and it is also more likely that the topic would trigger further discussions and create more hyperlinks. Thus, exponential exciting functions are well suited to capture such phenomenon as the exiting functions should have relatively large values at first and decay fast as time elapses.

For this experiment, we used the MemeTracker\footnote{\tt\scriptsize http://memetracker.org/data/links.html} dataset.
The data contains time-stamped phrase and hyperlink information for news media articles and blog posts from over a million different websites.
We extracted the times that hyperlinks to 10 well-known websites listed in Table \ref{table} are created during August 2008 to April 2009. 
When a hyperlink to a website is created at a certain time, an arrival events is recorded at that time. %The time step between records were one second. 
More precisely, in this experiment, we picked 30 different phrases that appeared on different websites at different times. 
If a website that published one of the phrases at time $t$ also contained a hyperlink to one of the 10 listed websites, an arrival event was recorded at time $t$ for that website in our list.

\begin{figure*}[t]
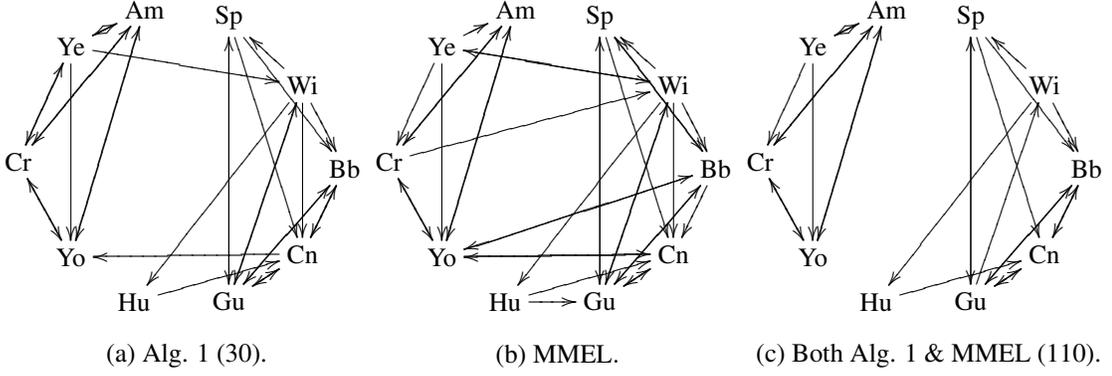

 \hspace{-.1cm}
\xygraph{ !{<0cm,0cm>;<1.4cm,-.3cm>:<-0cm,1.4cm>::} 
 !{(-.6,-.2) }*+{\text{Cr}}="x1"
!{(-.1,1) }*+{\text{Ye}}="x2"
 !{(.6,1.5) }*+{\text{Am}}="x3" 
 !{(1.4,1.6)}*+{\text{Sp}}="x4"
 !{(2.1,1.1) }*+{\text{Wi}}="x5"  
!{(2.5,.4) }*+{\text{Bb}}="x10"
!{(2.1,-.5) }*+{\text{Cn}}="x7" 
!{(1.4,-1.1) }*+{\text{Gu}}="x8" 
!{(.5,-1.3) }*+{\text{Hu}}="x9" 
!{(-.1,-1) }*+{\text{Yo}}="x6" 
!{(1,-1.7) }*+{\text{(a)\ Alg. \ref{algorithm} (30).}}
"x8":"x10" "x10":"x8"
 "x1":"x3" "x3":"x1"
  "x6":"x3" "x6":"x1"  "x1":"x6"   "x3":"x6"
   "x9":"x7" "x4":"x7"
     "x4":"x10" "x1":"x2"
    "x5":"x4" "x5":"x7" "x5":"x8" "x8":"x5" "x5":"x9" "x5":"x10" "x4":"x8" "x8":"x4"
    "x2":"x6" "x2":"x1"  "x2":"x3" "x3":"x2" "x7":"x10" "x10":"x7" "x7":"x8" "x8":"x7" "x2":"x5"
    "x7":"x6" 
}
\xygraph{ !{<0cm,0cm>;<1.4cm,-.3cm>:<-0cm,1.4cm>::} 
 !{(-.6,-.2) }*+{\text{Cr}}="x1"
!{(-.1,1) }*+{\text{Ye}}="x2"
 !{(.6,1.5) }*+{\text{Am}}="x3" 
 !{(1.4,1.6)}*+{\text{Sp}}="x4"
 !{(2.1,1.1) }*+{\text{Wi}}="x5"  
!{(2.5,.4) }*+{\text{Bb}}="x10"
!{(2.1,-.5) }*+{\text{Cn}}="x7" 
!{(1.4,-1.1) }*+{\text{Gu}}="x8" 
!{(.5,-1.3) }*+{\text{Hu}}="x9" 
!{(-.1,-1) }*+{\text{Yo}}="x6" 
!{(1,-1.7) }*+{\text{(b)\ MMEL}.}
"x8":"x10" "x10":"x8"
 "x1":"x3" "x3":"x1"
  "x6":"x3" "x6":"x1"  "x1":"x6"   "x3":"x6"
   "x9":"x7" "x4":"x7"
    "x10":"x4" "x4":"x10"
    "x5":"x4" "x5":"x7" "x5":"x8" "x8":"x5" "x5":"x9" "x5":"x10" "x4":"x8" "x8":"x4" 
    "x2":"x6" "x2":"x1" "x2":"x3" "x10":"x7" "x7":"x8" "x8":"x7" 
    "x2":"x5" "x5":"x2" "x9":"x8" "x7":"x6" "x6":"x7" "x10":"x6" "x6":"x10" "x1":"x5"
}
\xygraph{ !{<0cm,0cm>;<1.4cm,-.3cm>:<-0cm,1.4cm>::} 
 !{(-.6,-.2) }*+{\text{Cr}}="x1"
!{(-.1,1) }*+{\text{Ye}}="x2"
 !{(.6,1.5) }*+{\text{Am}}="x3" 
 !{(1.4,1.6)}*+{\text{Sp}}="x4"
 !{(2.1,1.1) }*+{\text{Wi}}="x5"  
!{(2.5,.4) }*+{\text{Bb}}="x10"
!{(2.1,-.5) }*+{\text{Cn}}="x7" 
!{(1.4,-1.1) }*+{\text{Gu}}="x8" 
!{(.5,-1.3) }*+{\text{Hu}}="x9" 
!{(-.1,-1) }*+{\text{Yo}}="x6" 
!{(1,-1.7) }*+{\text{(c)\ Both Alg. \ref{algorithm} \& MMEL (110).}}
"x8":"x10" "x10":"x8"
 "x1":"x3" "x3":"x1"
  "x6":"x3" "x6":"x1"  "x1":"x6"   "x3":"x6"
   "x9":"x7" "x4":"x7"
     "x4":"x10"
    "x5":"x4"   "x8":"x5" "x5":"x9" "x5":"x10" "x4":"x8" "x8":"x4"
    "x2":"x6" "x2":"x1"  "x2":"x3" "x3":"x2" "x7":"x10" "x10":"x7" "x7":"x8" "x8":"x7" 
}
\caption{Recovered causal structure of the MemeTracker dataset using (a) Algorithm \ref{algorithm}, (b) MMEL for $30$ different phrases, and (c) both Algorithm \ref{algorithm} and MMEL for $110$ different phrases.}  \label{meme}
\end{figure*}

Figure \ref{meme}(a) illustrates the resulting causal structure learned by Algorithm \ref{algorithm} for $z=12$ hours and $\Delta=1$ hour.
In this graph, an arrow from a node to another, say node \text{Ye} to \text{Yo}, means creating a hyperlink to \texttt{yelp.com} triggers creation of further hyperlinks to \texttt{youtube.com}.
% This is an interesting result that suggests the causal structure of a network can also be used for clustering the processes.

We also applied the MMEL algorithm with one exponential kernel function to learn the excitation matrix.
For this experiment, the data corresponding to each phrase was treated as an i.i.d. realization of the system.
The resulting causal structure is depicted in Figure \ref{meme}(b).

As Figure \ref{meme}(a) illustrates, the nodes can be clustered into two main groups: \{Cr, Ye, Am, Yo\} and \{Bb, Cn, Gu, Hu, Sp, Wi\}. The first group consists of mainly merchandise and reviewing websites and the second group contains the broadcasting websites. However, this is not as clear in Figure \ref{meme}(b). This is because MMEL requires more i.i.d. samples (phrases) to be able to identify the correct arrows.
Note that as we increase the number of phrases (110), Figure \ref{meme}(c), both graphs become similar with two clearly visible main clusters.

%Given that both methods use exponential for their exciting functions and the dataset is rich enough in terms of number of samples, it is not surprising that both approaches learn nearly the same excitation matrix.

\begin{center}
\begin{table}
\hspace{4.5cm}    
    \begin{tabular}{ | l  | p{3.5cm} |}
    \hline
   Cr &  \tt craigslist.org \\ \hline
    Ye & \tt yelp.com \\ \hline
     Am &  \tt amazon.com \\ \hline
     Sp &  \tt spiegel.de \\ \hline
     Wi &  \tt wikipedia.org \\ \hline
     Yo &  \tt youtube.com \\ \hline
     Cn & \tt cnn.com \\ \hline
     Gu & \tt guardian.co.uk \\ \hline
     Hu & \tt humanevents.com\\    \hline
      Bb & \tt bbc.co.uk\\ \hline
  \end{tabular}\caption{List of websites studied in MemeTracker experiment.}\label{table}
  \end{table}
  \end{center}
  
%===========================================================================
\section{Conclusion and Future Work}\label{sec:con}
Learning the causal structure (DIG) of a stochastic network of processes requires estimation of conditional directed information (\ref{cdif}). Estimating this quantity in general has high complexity and requires a large number of samples. However, the complexity of the learning task could be significantly reduced, if side information about the underlying structure of system dynamics is available. As proved in \ref{l1}, for multivariate Hawkes processes, estimating the support of the excitation matrix suffices to learn the associated DIG. Therefore, all approaches for learning the excitation matrix of the multivariate Hawkes processes such as ML estimation \cite{ML, ML2}, EM algorithm \cite{EM}, non-parametric estimation techniques proposed in \cite{bacry}, and the proposed method in this paper may  be used to learn the causal interactions in such networks. The previous estimation approaches either require i.i.d. samples such as MMEL or are limited to the class of symmetric Hawkes processes. The proposed algorithm in this work allows us to learn the support of the excitation matrix in a larger class of matrices in the absence of i.i.d. samples.

%\subsubsection*{Acknowledgements}

\section{Technical Proofs}
\subsection{Proof of Proposition \ref{l1}}\label{l1p}
Suppose $\gamma_{i,j}\equiv0$. (\ref{intensities}) implies that for every $t\leq T$, $\lambda_{i}(t)$ is $\underline{\mathcal{F}}^t_{-\{j\}}(=\sigma\{\underline{N}^{t}_{-\{j\}}\})$-measurable and from (\ref{prob}), we have
$$
\small{P\left(dN_{i}(t)=1|\underline{\mathcal{F}}^{t}\right)=P(dN_{i}(t)=1|\underline{\mathcal{F}}^{t}_{-\{j\}})}.
$$
Equivalently, for every $0\leq t_{k-1}<t_{k}$,
\begin{equation}\label{p1}
I\left(N_{i,t_{k-1}}^{t_k};N_{j,0}^{t_k}|\mathcal{F}^{t_{k-1}}_{-\{j\}}\right)=0,
\end{equation}
and thus, 
$\tilde{I}_{\textbf{t}}(N_{j}\rightarrow N_{i}||\small{\underline{N}}_{-\{i,j\}})=0$, for any finite partition $\textbf{t}\in\mathcal{T}(0,T)$.
\\
For the converse we use proof by contradiction. Suppose $I_T(N_{j}\rightarrow N_{i}||\underline{N}_{-\{i,j\}})=0$ and $\gamma_{i,j}\neq0$. 
Using the definition in (\ref{cdif}), it is straightforward to observe that for any $t<T$, 
$$
I_t(N_{j}\rightarrow N_{i}||\underline{N}_{-\{i,j\}})=0.
$$
Similarly, $I_{t+dt}(N_{j}\rightarrow N_{i}||\underline{N}_{-\{i,j\}})=0$. Consequently, 
$$
0=I_{t+dt}(N_{j}\rightarrow N_{i}||\underline{N}_{-\{i,j\}})-I_{t}(N_{j}\rightarrow N_{i}||\underline{N}_{-\{i,j\}})
$$
$$
=I\left(dN_{i}(t);N_{j,0}^{t}|\mathcal{F}^{t}_{-\{j\}}\right).
$$
This implies 
%Since, for any $\textbf{t}'$ that is a refinement of $\textbf{t}$, i.e., $\{t_i\}\subset\{t'_i\}$, $I_\textbf{t}'(N_{j}\rightarrow N_{i}||\small{\underline{N}}_{-\{i,j\}})\leq I_\textbf{t}(N_{j}\rightarrow N_{i}||\small{\underline{N}}_{-\{i,j\}})$ \cite{weissman2013directed}, we obtain there exists a small enough $dt$ such that
%any finite partition $\textbf{t}$, equation (\ref{p1}) holds and together with (\ref{prob}) imply that for every $t(=t_{k-1})$, and $t+dt(=t_k)$,
$
P(dN_{i}(t)=1|\underline{\mathcal{F}}^{t}_{-\{j\}})=\lambda_{i}(t)dt +o(dt),
$
or $\lambda_{i}(t)$ is $\underline{\mathcal{F}}^{t}_{-\{j\}}$-measurable. Since, we have assumed $\gamma_{i.j}\neq0$, we obtain $N_{j}(t)$ is $\underline{\mathcal{F}}^{t}_{-\{j\}}$-measurable, for all $t\leq T$. In words, $j$th process is determined by other processes which contradicts with the Assumption \ref{strictly} that states there is no deterministic relationships between processes.

%\subsection{Proof of Proposition \ref{l12}}\label{l12p}
%Suppose, there exists a function $G_i$ and an independent exogenous noise $\xi_i$, such that 
%\begin{equation}\label{eqii}\notag
%dN_i(t)=G_i(\underline{N}^{t},\xi_i(t+dt),t),
%\end{equation}
%where $\underline{N}^{t}=\{N_1^t,...,N_m^t\}$. 
%%Using equations (\ref{prob}), we have $G_i(\underline{N}^{t},\xi_i(t+dt),t)=N_i(t)+\eta_{i}(t+dt)$,
% where $G_i$ is a binary random variable, such that 
% $$
% P(G_i(\underline{N}^{t},\xi_i(t+dt),t)=1|\underline{\mathcal{F}}^t)=\lambda_i(t)dt+o(dt).
% $$
%Remind that $\underline{\mathcal{F}}^t=\sigma\{\underline{N}^t\}$.
% If $G_i$ is not a function of $N_j$, then $G_i$ is a deterministic value given $\{\underline{N}^t,\xi_i(t+dt)\}\setminus\{N_j(t')\}$ for any $t'\leq t$. Because, $\xi_i(t+dt)$ is independent of $\underline{N}^t$, we obtain that $G_i$ is conditionally independent of $N_j(t')$ given $\{\underline{N}^t\}\setminus\{N_j(t')\}$ for any $t'\leq t$.
% \\
%Using the positivity assumption, we obtain that $G_i$ is also conditionally independent of $N^t_j=\bigcup_{t'\leq t}\{N_j(t')\}$ given $\underline{\mathcal{F}}^t_{-\{j\}}$. In other words, $\lambda_i(t)$ is $\underline{\mathcal{F}}^t_{-\{j\}}$-measurable, which implies that $\gamma_{i,j}\equiv0$.
%\\
%For the converse, suppose $\gamma_{i,j}\equiv0$, then clearly, $G_i$ is $\underline{\mathcal{F}}^t_{-\{j\}}$-measurable.

\subsection{Proof of Corollary \ref{col}}\label{colp}
If the excitation matrix belongs to $\mathcal{E}\textit{xp}(m)$, from Equation (\ref{lapl}) we have
\begin{align}\nonumber
&\small{\left(I-\sum_{d=1}^{D}\frac{A^T_{d}}{j\omega+\beta_{d}}\right)diag(\Lambda)^{-1}\left(I-\sum_{d=1}^{D}\frac{A_{d}}{-j\omega+\beta_{d}}\right)}\\ \nonumber
&=\frac{4\sin^{2}{z\omega/2}}{\omega^{2}z}\mathcal{F}[\Sigma_z]^{-1}(\omega).
\end{align}
By evaluating the trace of the above equation, we obtain 
\begin{small}
\begin{align}\label{ew}
\hspace{-1mm}\sum_{i=1}^{m}\dfrac{|1-a_{i,i}|^{2}}{\lambda_{i}}+\sum_{i\neq j}\dfrac{|a_{i,j}|^{2}}{\lambda_{i}}=\frac{4\sin^{2}{z\omega/2}}{\omega^{2}z} Tr\mathcal{F}[\Sigma_z]^{-1}(\omega),
\end{align}
\end{small}
where $a_{i,j}=\sum_{d=1}^{D}\frac{a_{i,j}^{(d)}}{-j\omega+\beta_d}$, and $A_{d}=[a^{(d)}_{i,j}]$. To learn the entire set $\{\pm j\beta_d\}$, we have to show that there are no pole zero cancellations in  (\ref{ew}). That is, the nominator and denominator of (\ref{ew}) have no common roots. Let  
$$
g(\omega):=\left(\sum_{i=1}^{m}\dfrac{|1-a_{i,i}|^{2}}{\lambda_{i}}+\sum_{i\neq j}\dfrac{|a_{i,j}|^{2}}{\lambda_{i}}\right)\prod_{d=1}^{D}|-j\omega+\beta_d|^{2},
$$
which is the nominator of Equation (\ref{ew}). It is straightforward to check that for $\omega=-j\beta_k$,
%$$
%g(-j\beta_{k})=\left(\sum_{i,j}\frac{|a^{(k)}_{i,j}|^{2}}{\lambda_i}\right)\prod_{i\neq k}(\beta_k-\beta_i)^2.
%$$
the above quantity is non-zero, due to the fact that $\beta_d$s are distinct and $A_k\neq\textbf{0}$. Since $g(\omega)$ is a polynomial with real coefficients, from  complex conjugate root theorem \cite{jeffrey2005complex}, we have  $g(j\beta_k)\neq0$. Therefore, the set $\{\pm j\beta_d\}$ contains all the poles of (\ref{ew}).

\subsection{Proof of Proposition \ref{pro}}\label{prop}
From Lemma \ref{l22}, the Laplace transform of the covariance density can be written as
\begin{align}\label{lapli}\nonumber
&\mathcal{L}[\Omega](s)=\mathcal{L}[\Gamma](s)\left(diag(\Lambda)+\mathcal{L}[\Omega](s)\right)\\ \nonumber
&+\int_{0}^{\infty}\int_{t}^{\infty}\Gamma(t')\Omega^{T}(t)e^{-s(t'-t)}dt'dt.
\end{align}
When $\Gamma(t)\in\mathcal{E}\textit{xp}(m)$, it can be shown that (\ref{lapli}) becomes
\begin{equation}\label{dasgha}
\mathcal{L}[\Omega](s)=\sum_{d=1}^{D}\frac{A_{d}}{s+\beta_{d}}\left(diag(\Lambda)+\mathcal{L}[\Omega](s)+\mathcal{L}[\Omega]^{T}(\beta_{d})\right).
\end{equation}
If the set of exciting modes are given, we can insert $s=\beta_{d}$, for $d=1,\dots,D$ in the above equation and obtain the system of $D$ equations. 

%\newpage

\bibliographystyle{plain}
\bibliography{ref}

\end{document}